# STRUCTURAL HEALTH MONITORING OF CANTILEVER BEAM, A CASE STUDY - USING BAYESIAN NEURAL NETWORK AND DEEP LEARNING

Rahul Vashisht[*], H.Viji[*,#], T.Sundararajan[#], D.Mohankumar[#], S.Sumitra[*]
(rahulvashisht290@gmail.com, viji.narayan@gmail.com, d_mohankumar@vssc.gov.in, t_sundararajan@vssc.gov.in, sumitra@iist.ac.in)
([*]Indian Institute of Space Science & Technology, Thiruvananthapuram, PIN – 695547, India
[#]Vikram Sarabhai Space Centre, ISRO, Thiruvananthapuram, PIN – 695022, India)

**ABSTRACT**

The advancement of machine learning algorithms has opened a wide scope for vibration based SHM (Structural Health Monitoring). Vibration based SHM is based on the fact that damage will alter the dynamic properties viz., strucural response, frequencies, mode shapes, etc of the structure. The responses measured using sensors, which are high dimensional in nature,can be intelligently analysed using machine learning techniques for damage assessment.Neural networks employing multilayer architectures are expressive models capable of capturing complex relationships between input-output pairs, but do not account for uncertainty in network outputs. A BNN (Bayesian Neural Network) refers to extending standard networks with posterior inference. It is a neural network with a prior distribution on its weights. Deep learning architectures like CNN (Convolutional neural network) and LSTM(Long Short Term Memory) are good candidates for representation learning from high dimensional data. The advantage of using CNN over multi layer neural networks is that they are good feature extractors as well as classifiers, which eliminates the need for generating hand engineered features. LSTM networks are mainly used for sequence modeling. This paper presents both a Bayesian multi layer perceptron and deep learning based approach for damage detection and location identification in beam-like structures. Raw frequency response data simulated using finite element analysis is fed as input of the network. As part of this, frequency response was generated for a series of simulations in the cantilever beam involving different damage scenarios (at different location and different extent). These frequency responses can be studied without any loss of information, as no manual feature engineering is involved. The results obtained from the models are highly encouraging. This case study shows the effectiveness of the above approaches to predict bending rigidity with an acceptable error rate.

**Keywords:** Bayesian neural network, Deep learning

## 1.0 INTRODUCTION

Deep learning models are being used on a daily basis to solve different tasks in vision, linguistics and signal processing[1-5]. Understanding whether the model is under-confident or falsely over-confident can help get better performance out of the model. Recognizing that test data is far from training data, one could easily augment the training data accordingly. In deep neural networks, usage of softmax to get probabilities is actually not enough to obtain model uncertainty. Standard neural network, with probability distribution over each of its weights is called BNN [6-11]. BNN gives the uncertainty estimates over the network outputs and can also help in model selection. PBP (Probabilistic backpropagation) is the learning technique used to train BNN in place of standard backpropagation.

This paper presents both a Bayesian multilayer perceptron based approach and deep learning based approach (using architectures viz., CNN & LSTM) for damage detection and location identification in cantilever beam. Raw frequency response data simulated using finite element analysis is fed as input of the network. Conventional data driven approaches make use of statistical techniques for feature extraction from raw signals, in which case, data transformation has to be done accordingly for new data. The present approach models on raw data, which makes it ideal for real time monitoring as it eliminates the need for data transformation. Details of each of the approaches are given in the following sections.

## 2.0 EXPERIMENTAL
### 2.1 Bayesian Neural Networks
Bayesian modelling is a powerful method to investigate the uncertainty in deep learning models[4].Bayes theorem tells us about how to do inference from data. It follows the rule below to calculate conditional distribution of hypothesis given observed data.

$$P(hypothesis \mid data) = \frac{P(data \mid hypothesis) * P(hypothesis)}{P(data)}$$

Thus learning and predictions can be seen as inference problems. Here P ( hypothesis) represents the prior beliefs over the hypothesis and P(data | hypothesis) represents likelihood of data given hypothesis which can be obtained from observed data.P( data) can be obtained by marginalising over the space of hypothesis and thus is considered as normalization constant with respect to the hypothesis parameters.
Given training inputs $\mathbf{X} = \{x_1, x_2, \ldots, x_N\}$ and their corresponding outputs $\mathbf{Y} = \{y_1, y_2, \ldots, y_N\}$, in Bayesian (parametric ) regression we would like to find the parameters $\omega$ of a function $y = f^\omega(x)$ that are likely to have generated the outputs. Following the Bayesian approach, some prior distribution is put over the space of parameters, $p(\omega)$. Prior distribution represents our belief as to which parameters are likely to have generated the data before we observe any data points. Once some data is observed the prior can be combined with the likelihood of the observed data to obtain the posterior which can be used for prediction of new data. The likelihood distribution $p(y \vee x, \omega)$, generates the output given the input and parameters $\omega$. For regression, Gaussian likelihood is assumed which is as follows :

$$p(y \vee x, \omega) = N(y; f^\omega(x), \tau^{-1} I) \tag{1}$$

where $\tau$ is the model precision,which corresponds to adding observation noise to model output with variance $\tau^{-1}$.

Using Bayes theorem, we can define the posterior as follows:

$$p(\omega \mid X, Y) = \frac{p(y \mid x, \omega) p(\omega)}{p(Y, X)} \tag{2}$$

This distribution captures the most probable function parameters given our observed data . For prediction of new input $x^¿$, the posterior distriburion and likelihood of new data $x^¿$ are combined and the parameters can be marginalised to obtain the predictive distribution for the new input, which is known as inference.

$$p(y^¿ \vee x^¿, X, Y) = \int_\omega p(y^¿ \vee x^¿, \omega) p(\omega \vee X, Y) d\omega \tag{3}$$

For most of deep learning models the posterior distribution is intractable. Hence posterior will be approximated by some approximating distribution $q(\omega \vee X, Y)$, which is obtained by minimizing the KL-divergence between the posterior distribution and the approximating distribution [6].

Applying the bayesian modelling gives the probabilistic interpretation of deep learning models by inferring distribution over model weights.Standard neural networks lacks the capability of uncertainty measurement, as it only offer point estimates. BNN offers robustness to overfitting, uncertainty estimates and can easily learn from small datasets.We have used the probabilistic backpropagation to train the neural network which is discussed in the next section.

### 2.2 Probabilistic Back Propagation
Probabilistic back propagation[8](technique to train BNN) works similar to standard back propagation (which is a technique to train a standard neural network). PBP uses a collection of one dimensional Gaussians in place of point estimates for the weights, each one approximating the marginal posterior distribution of all the weights in the respective layer. In the forward pass, input data is propagated forward through the network. Since the weights are random, the activation produced in each layer are random, and result in intractable distributions. It sequentially approximates each of these distributions with the collection of one dimensional Gaussians that match their marginal means and variances. At the end of this phase PBP computes the logarithm of the marginal probability of the target variable instead of prediction error.

$$m^{a_l} = M_l m^{z_{l-1}} / \sqrt{n_{l-1}+1}$$

$$v^{a_l} = [(M_l \circ M_l) v^{z_{l-1}} + V_l (m^{z_{l-1}} \circ m^{z_{l-1}}) + V_l v^{z_{l-1}}] / (n_{l-1}+1)$$

where $M_l$ and $V_l$ are $n_l$ x $n_{l-1}+1$ matrices whose entries are given by mean and variance of weight distribution. In the backward pass, the gradients of marginal distribution, Z is propagated backwards using reverse mode differentiation as in classical back propagation. These deriavtives are then, used to update the means and variances of the posterior approximation.

The normalization constant Z of posterior is obtained using approximation of integral as a normal distribution of the following form:

$$Z = N(y_n \vee m^{z_l}, vfinal)$$

where $y_n$ is the actual output and $m^{z_l}$ and is mean obtained after the forward pass, *vfinal* is variance obtained after forward pass and incorporating the prior information.

## 2.3 CNN

CNN model can be described as the recursive appplication of convolution and pooling layer, followed by inner product layer, commonly called as dense layers, at the end of the network. Convolutional layer, which is the core layer of CNN is a linear transformation that preserves the spatial information of the input space. Covolutional layer can be considered as a feature extraction layer.Pooling layers (optional) simply takes the output of the convolutional layer and reduce its dimensionality. Intuitively, it is capable of bringing shift and distortion invariance to the input and to avoid overfitting to some extent. A convolutional layer's output shape depends on the shape of its input , shape of kernel, zero padding and strides. The non linearity is incorporated in the network by applying activation functions such as relu,tanh,sigmoid etc after the convolution layer.A simple CNN, also known as ConvNet is a sequence of layers. Every layer of a ConvNet transforms one volume of activations to another through a differentiable function. Convolutional layer does the convolution operation, which computes the output of neurons that are connected to local regions in the input. The pooling layer operates independently on every depth slice of the input and resizes it spatially.The layers used to build ConvNets are detailed below:

<u>Convolutional Layer</u>

This layer is the core building block of CNN. This layer does the convolution operation,which computes the output of neurons that are connected to local regions in the input.The CONV layer's parameters consist of a set of learnable filters. Every filter(also known as kernel) is small spatially (along width and height), but extends through the full depth of the input volume. During the forward pass, each of the filters is convolved with the input volume and dot products are computed between the entries of the filter and the input at any position. For the convolution layer, there are four hyper-parameters viz., no.of filters K, the spatial extent of the filter F, the stride S and the amount of zero padding P.
An input volume of size W1 x H1 x D1 produces a volume of size W2 x H2x D2 , where
$$W_2 = (W_1 - F + P)/S + 1$$
$$H_2 = (H_1 - F + P)/S + 1$$
$$D_2 = K$$

The backward pass for a convolution operation is also a convolution (but withspatially-flipped filters)

Pooling Layer

The pooling layer operates independently on every depth slice of the input and resizes it spatially.The depth dimension remains unchanged. The pooling options available are maximum pooling, average pooling, L2-norm pooling etc. More generally, the pooling layer accepts a volume of size W1 x H1 x D1. The hyperparameters of this layer are their spatial extent F and the stride S. The resulting operation produces a volume of size W2 x H2 x D2 where

$$W_2 = (W_1 - F)/S + 1$$
$$H_2 = (H_1 - F)/S + 1$$
$$D_2 = D_1$$

The backward pass for a max(x,y) operation has a simple interpretation as only routing the gradient to the input that had the highest value in the forward pass.This layer is not mandatory and in some cases, they are discarded.

Fully-connected layer

Neurons in a fully connected layer have full connections to all activations in the previous layer, as seen in standard neural networks.The backward pass for FC layer can be done similar to CONV layer, where the spatial extent of the filteris the same as that of input volume.

Unlike traditional filters that have predefined parameters, the parameters of the 2D filter kernels in CNNs are automatically optimized by back propagation.

## 2.4 LSTM

LSTM are a special kind of RNN (Recurrent Neural Network), capable of learning long-term dependencies [15]. They were introduced by Hochreiter and Schmidhuber. The core idea behind LSTMs lies that at each time step, a few gates are used to control the passing of information along the sequences that can capture long-range dependencies more accurately. LSTM have a chain of repeating modules of neural network, each having four neural network layers. The key to LSTMs is the cell state. The cell state runs straight down the entire chain, with only some minor linear interactions. LSTM consists of stuctures called gates, through which information is added or removed to the cell state. Gates are a way to optionally let the information through, composed out of a sigmoid neural network layer and a pointwise multiplication operation. At each time step t, hidden state $h_t$ is updated by current data at the same time step, hidden state at previous time step, input gate, forget gate, output gate and a memory cell. The decision regarding what information is to be through away from cell state is made by the forget gate layer. The forget gate accepts input at the current time step and hidden state at previous time step and outputs a number between 0 and 1 for each number in the cell state $c_{t-1}$. If the value is 0, it means to completely get rid of the information, while 1 means to completely retain.

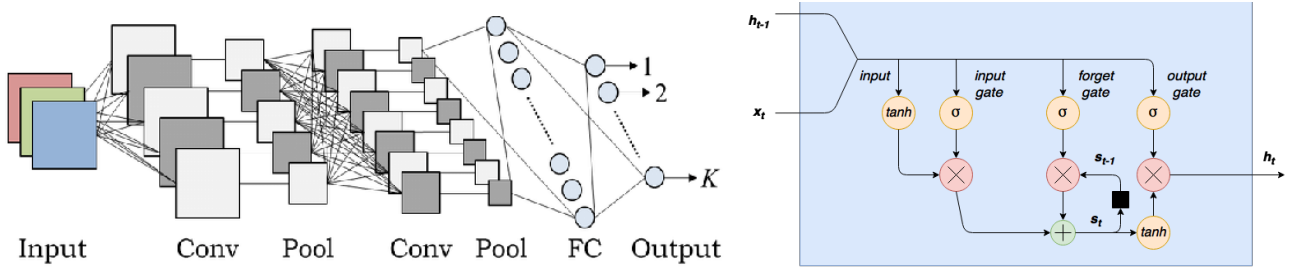

Figure 1: Sample CNN[21] (left) and LSTM[22] architecture(right)

The basic LSTM equations are:

$$a_t = \tanh(W_a x_t + U_a h_{t-1} + b_a)$$

$$i_t = \sigma(W_i x_t + U_i h_{t-1} + b_i)$$
$$f_t = \sigma(W_f x_t + U_f h_{t-1} + b_f)$$
$$o_t = \sigma(W_o x_t + U_o h_{t-1} + b_o)$$
$$c_t = f_t c_{t-1} + i_t \tanh(W_c x_t + U_c h_{t-1} + b_c)$$
$$h_t = o_t \tanh(c_t)$$

where model parameters including all $W \in R^{dxk}$, $U \in R^{dxd}$ and $b \in R^d$ are shared by all time steps and learned during model training, σ is the sigmoid activation function, denoting the element-wise product, k is a hyper-parameter that represents the dimensionality of hidden vectors. The basic LSTM is constructed to process the sequential data in time order and the output at the terminal time stepis used to predict the output.

## 2.5 Case Study

A cantilever beam structure of circular cross section having a length of 1m is considered for the study. The structure is divided into four beam elements as shown in Figure 2, where N stands for node and E stands for element[16-18]. Each element is having initial diameter of 0.01m. Euler-Bernoulli beam element is used to idealise the structure. The nodal variables are transverse deflection and slope.

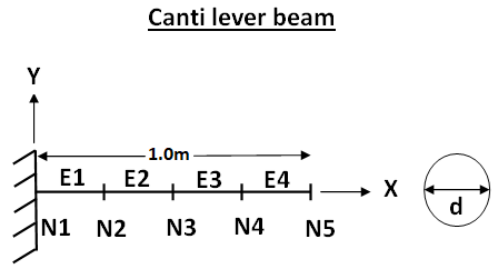

Figure 2: Cantilever beam

where EI is the beam rigidity and *l* is the length of the element. The stiffness and mass matrix of the beam element is as follows:

$$K = EI/l^3 \begin{bmatrix} 12 & 6l & -12 & 6l \\ 6l & 4l^2 & -6l & 2l^2 \\ -12 & -6l & 12 & -6l \\ 6l & 2l^2 & -6l & 4l^2 \end{bmatrix} \quad M = \rho Al/420 \begin{bmatrix} 156 & 22l & 54 & -13l \\ 22l & 4l^2 & 13l & -3l^2 \\ 54 & 13l & 156 & -22l \\ -13l & -3l^2 & -22l & 4l^2 \end{bmatrix}$$

where $\rho$ is the mass density, A is the area and *l* is the length of the element. Damage is simulated by changing the diameter of the individual elements. Frequency response analysis is carried out from 0.1rad/sec to 10000rad/sec for 1N excitation by varying the diameter of the beam element from 0.005m to 0.015m. The excitation force is applied at node number 5 in Y direction. The acceleration response is monitored at node numbers 2, 3, 4 and 5 in the Y direction.

## 3.0 RESULTS AND DISCUSSION

Response from each node is of length 10,000. Here, a single data point is obtained by concatenating responses in the order 5,4,3,2 forming a vector of length 40,000. Sample input is shown in Figure 3, where x-axis represents four responses (where each omega varying from 1 to 10000) and y-axis corresponds to normalised acceleration. The output for each data point is a vector of length 4 corresponding to four diameter of the elements E1, E2, E3 and E4. Total input data points are 14,641.
For training model using deep learning architectures, data was divided into training-testing with a split of 70:30. Training data was further split into training-validation in the ration 70:30. Adam was used as the optimiser. Mean squared error was used as the loss function.

### 3.1 CNN model architecture

To make the input data compatible for applying 2D CNN, each data point of length 40,000 is reshaped to 200 x 200. To predict diameter for each element, four neural network models were developed. The number of convolution and pooling layers used in each network with the respective hyperparameters are given in Table 1. After each convolution layer, RELU non-linearity is applied. Only for the first convolution, zero-padding is done for all models. The last two layers for all models are Global average pooling and Dense layer (with one neuron). The R-squared score for this model is 0.9862.

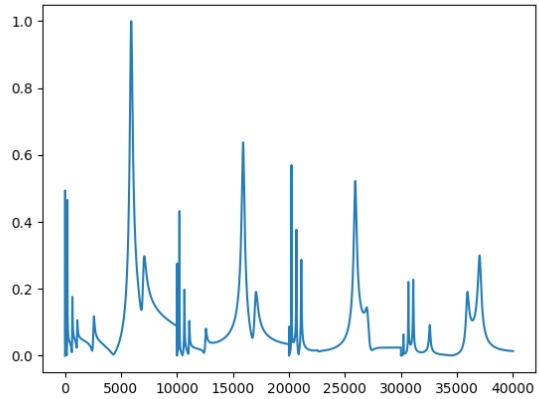

Figure 3: Sample input of cantilever beam study. frequency vs acceleration

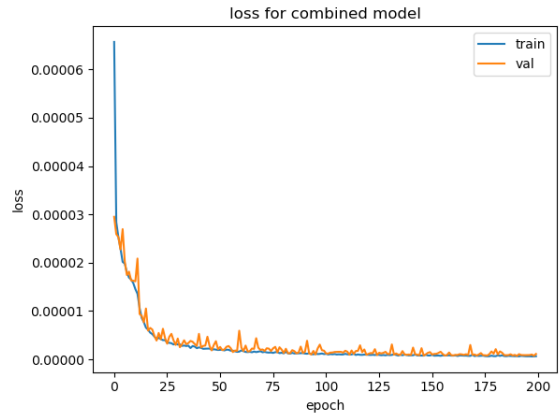

Figure 4: 2D CNN Loss curve for cantilever beam

Table 1: Hyperparameters used in 2D CNN architecture for each beam element

|            | Conv1    | MaxPool1 | Conv2   | MaxPool2 | Conv3   | MaxPool3 | Conv4  |
|------------|----------|----------|---------|----------|---------|----------|--------|
| *Element E1* | 32 (3,3) | (4)      | 32 (3,3)| (2)      | 32 (3,3)| (2)      | 8 (3,3)|
| *Element E2* | 64 (5,5) | (4)      | 32 (3,3)| (2)      | 32 (3,3)| -        | -      |
| *Element E3* | 64 (5,5) | (4) | 32 (3,3) | (2) | 16 (3,3) | - | - |
| *Element E4* | 16 (3,3) | (2) | 32 (3,3) | (2) | - | - | - |

## 3.2 LSTM model architecture

To make the input data compatible for applying LSTM, each data point of length 40000 is reshaped to (4, 10000), where 4 represents the number of time steps. The data is modelled using a stack of three LSTM layers with 32, 16 and 4 hidden neurons respectively. The loss curve is given in Figure 5.The R-squared score for this model is 0.9910.

## 3.3 PBP model architecture

We have used the multi layer perceptron architecture with single hidden layer and 64 hidden units. It is trained using probabilistic back propagation for 10 epoch with train test split of 50.The R-squared score for this model is 0. 9967.

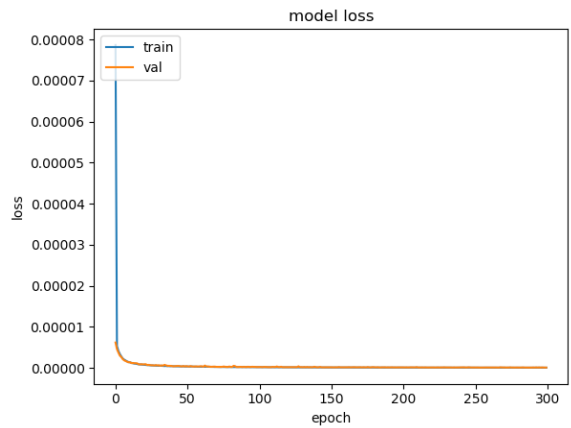

Figure 5: LSTM validation loss for cantilever beam

## 3.4 Model evaluation

To evaluate the efficiency of the model, R-squared score is used as the metric. The R-squared metric providesan indication of the goodness of fit of a set of predictions to the actual values. Itis a statistical measure that measures how close the data are to predictions.$R^2=1-SSE/SST$, where SSE is the sum of squares of difference between actual values and predicted values and SST is the sum of squares of difference between actual values and mean value.

R-squared statistic is defined as the percentage of the response variablevariation that is explained by a linear model. It is always between 0 and 1. Herewe want our predicted values to be as close as possible to the actual values, henceR-squared statistic should be close to 1. All the three models have given R-squared score, close to 1, which indicates the effectiveness of these algorithms to be applied on vibration based SHM.

Table 2: R squared score of all architecures for cantilever beam

| Algorithm | Test score |
|---|---|
| PBP | 0.9967 |
| LSTM | 0.9910 |
| 2D-CNN | 0.9862 |

## 3.5 Visualisation using CNN model

In regression outputs, one could visualize attention over input that increases, decreases or maintains the regressed filter index output. In this case study, this only tells us parts of the input that contribute towards to increase, decrease or maintaining the output value[13]. By default, ActivationMaximization loss yields positive gradients for inputs regions that increase the output and negative gradients that decrease the output. To visualize what contributed to the predicted output, gradients that have very low positive or negative values have to be considered. This can be achieved by performing grads = abs(1 / grads) to magnifies small gradients.

A random test case taken for the study is shown in Fig 5. The actual vs predicted output for the same is given in Table 11.

Table 11: Table showing predicted and actual values for a test case of cantilever beam

|  | E1 | E2 | E3 | E4 |
| --- | --- | --- | --- | --- |
| Actual diameter | 0.011 | 0.008 | 0.015 | 0.015 |
| Predicted diameter | 0.01130249 | 0.00846803 | 0.01471727 | 0.01560052 |

For our case study we have take top 10 gradients, which are shown in Figures 6 and 7. The green lines shows the top 10 features that maintain/contribute to the prediction. The blue lines shows those features that can contribute to increase in the predicted value and orange lines are the ones that can contribute to decrease in the predicted value.

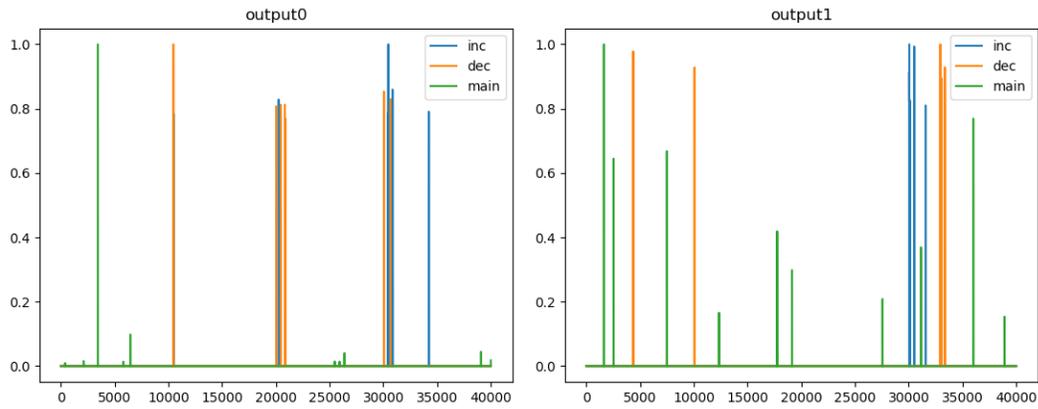

Figure 6: Visualisation for outputs E1 and E2

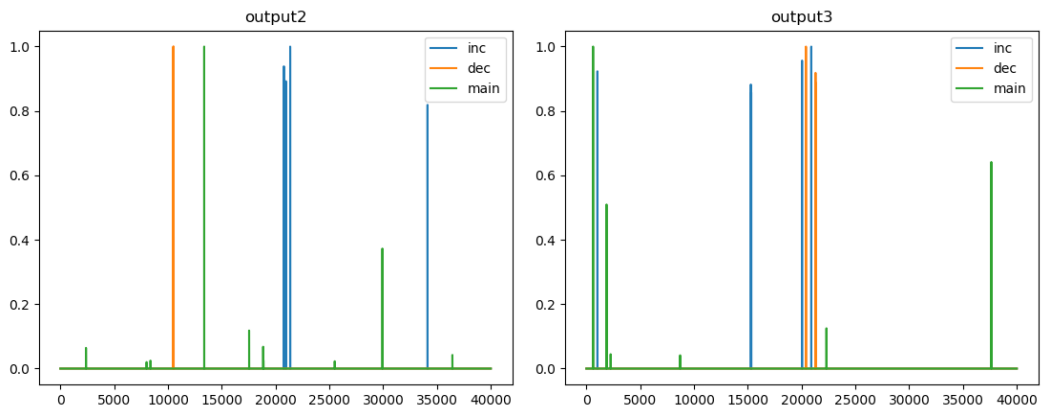

Figure 7: Visualisation for outputs E3 and E4

By visualising the gradients, it is evident that the first few modes are only required for identifying the damage near the fixed end where as higher modes are required for identifying the damage near the free end, which is clearly reflected in gradient visualisation.

## 4.0 CONCLUSIONS

The R-squared score obtained from all the three models are close to 1 for the unseen data, which shows that predictions are close to actuals and hence their generalisation capability. The R-squared score reflected by Bayesian model appears to outperform all the models. Moreover, gradient visualisation has demonstrated that neural networks are not just black boxes. It also gives us an inituition of the learning process, which closely associates with the underlying theory. The capability of deep learning architectures has shown their feature engineering capability, which could provide good predictions from raw vibration signals. This makes it ideal for real time monitoring as it eliminates the need for data transformation.

## ACKNOWLEDGEMENTS

The authors would like to acknowledge Director, VSSC and Director, IIST for permitting us to carry out this study.